# Identifying epidemic related Tweets using noisy learning

Ramya Tekumalla and Juan M. Banda
Department of Computer Science, Georgia State University
{rtekumalla1, jbanda} @gsu.edu

**Abstract**

Supervised learning algorithms are heavily reliant on annotated datasets to train machine learning models. However, the curation of the annotated datasets is laborious and time consuming due to the manual effort involved and has become a huge bottleneck in supervised learning. In this work, we apply the theory of noisy learning to generate weak supervision signals instead of manual annotation. We curate a noisy labeled dataset using a labeling heuristic to identify epidemic related tweets. We evaluated the performance using a large epidemic corpus and our results demonstrate that models trained with noisy data in a class imbalanced and multi-classification weak supervision setting achieved performance greater than 90%.

## 1. Introduction

Social media is where people digitally converge during disasters and use it as a lifeline for communication during natural disasters, epidemics, war and other crises. There have been over 1.3 billion Covid-19 tweets retrieved from the 1% sample of the Twitter data over a period of 2 years (Banda et al. 2021). This demonstrates that people tend to heavily rely on Twitter for communication during epidemics and additionally displays that Twitter contains an abundance of data signals which can be used for research.

Several studies in the past demonstrated successful results using NLP (Pawelek et al. 2014; Khatua et al. 2019) and supervised learning techniques (Lampos et al. 2010; Aramaki et al. 2011) for epidemics research. However, in recent times there has been a shift in relying towards other forms of machine learning techniques to avoid the manual curation process involved in supervised learning to create a ground truth dataset. In this aspect, **weak supervision** utilizes noisy, limited, or imprecise sources to provide a supervision signal for labeling large amounts of training data in a supervised learning setting (Ratner et al. 2019). By adopting heuristics and defining labeling functions which utilize the heuristics to label the dataset, large training sets i.e silver standard data could be programmatically generated and can be utilized for training machine learning models. Our work is based on the theory of noisy learning (Aslam and Decatur 1996; Simon 1996). Since the silver standard dataset is not manually verified it contains noise and hence models trained with the silver standard dataset are weak models as compared to models trained with gold standard.

In this work, we collected over 7 billion tweets from Twitter between 2013 and 2021. We filtered 8 different types of epidemic tweets using a heuristic approach and curated a silver standard dataset. We trained several machine learning models using the silver standard dataset and validated the performance of the models using a large epidemic corpus (Liu et al. 2020) which contains over 30 million epidemic tweets. To the best of our knowledge, we are the first to utilize weak supervision techniques for epidemics research.

## 2. Data Collection from Twitter

We utilized the Social Media Mining Toolkit (SMMT) (Tekumalla and Banda 2020) and acquired data in two different approaches. We utilized the "*get_metadata*" utility to hydrate 34 publicly available datasets and "*streaming*" utility to collect the 1% sample data from the Twitter end point which is an ongoing longitudinal data collection which started in 2018. The 34 publicly available datasets are listed in Appendix due to limitation of space. We collected a total of **7,301,497,397 tweets** from publicly available datasets and Twitter Stream.

### 2.1 Heuristic Creation for labeling the data

To create a heuristic, we first identified all the epidemics that occurred between 2006 and 2019. 2006 was our starting point since Twitter was established in 2006. We intentionally did not include Covid-19 as there are several large datasets on Covid-19 (Chen et al. 2020; Banda et al. 2021; Lamsal 2021) and very limited datasets on other epidemics. We identified 8 different deadly epidemics which are Cholera, Ebola, H1N1, HIV, Influenza, MERS, SARS, Yellow Fever. In addition to the epidemics, we also identified virus variants for few epidemics (Eg: Swine flu is a virus variant of H1N1; AIDS is caused by HIV). In this work, we used regular expressions as our labeling heuristic since for epidemics like "cholera", we wanted to retrieve all the tweets irrespective of case. To summarize, for epidemics Cholera, Ebola, H1N1, Influenza, flu, HIV, MERS and SARS we used expressions which would filter tweets irrespective of case sensitivity. For AIDS, the expression would only filter tweets if the term AIDS is in upper case in the tweet text.

## 2.2 Silver Standard Dataset Creation

To create the silver standard dataset, we filtered 8 different types of English epidemic tweets using a heuristic approach and filtered tweets from both publicly available datasets and twitter streams. We removed duplicate tweets and preprocessed the tweet text by removing emojis, emoticons, URLs and striped white spaces. Table 1 contains the details of the data collection and filtered tweets. In this work, we only utilize the text of the tweet and do not consider emoticons to add value to the text. The original tweets in table 1 are original tweets from the dataset while the Total tweets represent a total of both original and retweets. We only used original tweets to avoid adding bias to the model. While we collected over 7 billion tweets and filtered over 2 million tweets, after removing the duplicates, our silver standard dataset contains **2,302,924 tweets.** Table 2 presents the number of tweets filtered for each epidemic. The epidemic tweets in bold text are the tweets used in this work. Unsurprisingly flu has the most number of tweets (58.2%) of the epidemic tweets since it is more prevalent than other epidemics. 19.2% of the filtered tweets are from the epidemic Ebola. We intentionally separated flu and influenza tweets since flu is more prevalent in Internet language than influenza.

*Table 1*. Data collection details

| Data Source | Total Tweets | Original Tweets | Filtered Tweets |
|---|---|---|---|
| 34 Publicly available datasets | 3,050,058,283 | 765,974,491 | 2,095,057 |
| Twitter Stream (2018 - 2021) | 4,251,439,114 | 2,129,383,609 | 325,125 |
| **Total** | **7,301,497,397** | **2,895,358,100** | **2,420,182** |

## 3. Methods

To test the weak supervision approach, we trained several machine learning models using the silver standard. We used the filtered tweets from the **Cholera, Ebola, MERS and Swine Flu** and labeled each class separately. We collected an equal number of non epidemic tweets i.e tweets that do not contain any of the epidemics in the tweet text and labeled them as non epidemic samples. We utilized a stratified ratio of 75-25 of the dataset as training and validation data. For testing the models, we utilized a publicly available large epidemic copus (Liu et al. 2020) collected from the Twitter stream between 2009 and 2020. The corpus contains 30 million tweets from 4 epidemics (Cholera, Ebola, MERS and Swine Flu) in several languages. We could hydrate only 27,903,463 tweets from the Epic corpus as tweets were not available since they were either removed or deleted. Out of the 18,367,000 filtered tweets, 4,548,519 tweets are Swine Flu tweets, 1,020,094 tweets are Cholera tweets, 11,092,583 tweets are Ebola tweets and 133,011 are MERS tweets. This is the largest publicly available non-covid19 epidemic corpus. However, this corpus was collected based on keywords from Twitter stream. We labeled each class separately and to this we added an equal number of non epidemic tweets and labeled them as non epidemic samples. To summarize, our highly imbalanced silver standard data was utilized as training data which contains a total of **1,090,374** samples from 5 different classes where the number of negative tweets are greater than individual epidemic tweets. The highly imbalanced EPIC corpus was as test data which contains 33,588,414 samples from 5 different classes.

*Table 2*. Counts of Epidemic tweets in silver standard dataset

| Epidemic | No of filtered tweets |
|---|---|
| **Cholera** | **18,375** |
| **Ebola** | **441,035** |
| Flu | 1,340,557 |
| H1N1 | 100,146 |
| HIV/AIDS | 200,291 |
| Influenza | 410,60 |
| **MERS** | **8,993** |
| SARS | 66,980 |
| **Swine Flu** | **76,784** |
| Yellow Fever | 8,703 |

## 3.1 Machine Learning Models

We experimented with three classical models which include SVM, Decision Tree, and Logistic Regression models using the Scikit learn (Pedregosa et al. 2011) python library and 2 different Transformer models which include BERT (B) (Devlin et al. 2018), and BERTweet (BT) (Nguyen et al. 2020). Due to limitation of space, the model specifications are detailed in Appendix.

## 4. Results

To evaluate the performance of the models, we used four different metrics, Precision (P), Recall (R), F-Measure (F) for each class and also calculated Accuracy (A). While we obtained the performance on 4 metrics, we present only the results of F-measure in this work due to space limitation. The results for other metrics are presented in the Appendix. Table 3 presents the results of F-measure for all the 5 machine learning models.

*Table 3*. F-measure of the machine learning models

| Epidemic class | Classical Models | | | Deep Learning | |
|---|---|---|---|---|---|
| | DT | LR | SVM | B | BT |
| Cholera | 0.1826 | 0.9816 | **0.9836** | 0.9834 | **0.9822** |
| Ebola | 0.474 | 0.9736 | **0.9797** | 0.9925 | **0.9924** |
| MERS | 0.2673 | 0.9346 | **0.9337** | 0.8947 | **0.874** |
| Swine Flu | 0.1843 | 0.8441 | **0.8598** | 0.8231 | **0.8163** |
| non epidemic | 0.7244 | 0.9723 | **0.9726** | 0.9592 | **0.9576** |
| weighted (5 classes) | 0.5503 | 0.9555 | **0.9599** | 0.9455 | **0.9504** |

In all our experiments, all the models except the decision tree model, were able to accurately classify different classes of epidemic tweets using the noisy silver standard dataset despite the nature of the dataset being heavily imbalanced. The Ebola class had the highest number of samples both in train and the test set while the other samples were significantly lower. The non epidemic class perhaps has the most number of samples as the samples are equal to the sum of all the epidemic samples in the set. All the models consistently demonstrated a low performance in classifying the swine flu tweets when compared to other classes. Additionally, we calculated weighted F-measure for all the models and all the models had a performance of 95% while the decision tree had a score of 55%. The deep learning models performed as good as the classical models and in some classes (Eg: Ebola) performed better than the classical models. All the models, except the decision tree had a performance greater than 90%.

Our previous work utilizing weak supervision in a pharmacovigilance application (Tekumalla and Banda 2021) demonstrated high performance on the test set. However, this is the first application utilizing a multi-classification approach using weak supervision in this domain. In this work, we demonstrate that we could successfully classify different epidemics from a large scale epidemic corpus using silver standard dataset and machine learning models. While the EPIC corpus is not a gold standard corpus, it is the only largest publicly available "multi-class" epidemic corpus. While the EPIC corpus was a targeted collection, we present an approach which can easily extend to future epidemics and obtain noisy silver standard datasets for training machine learning models. We believe that when utilizing weak supervision methodology, we need to experiment with multiple models as there are no proven models that work for all applications. The important observation to note is that, it is easier to generate a silver standard dataset than manually curating a training set for supervised learning. Our results in this work validate our approach as we used a very large corpus for evaluation instead of a small dataset.

## 5. Future Work

In this work, we utilized a heuristic based on regular expressions of epidemics to label and create the silver standard data. In future, we would like to experiment with labeling functions and apply weak supervision frameworks like Snorkel (Ratner et al. 2020) and SkWeak (Lison et al. 2021) to generate large multi class training data. Additionally, we would like to train a BERT transformer model on all the collected epidemic tweets and release the Epidemic-BERT to the research community.

## 6. Conclusion

In this work, we curated a silver standard dataset of 8 different types of epidemics. We collected over 7 billion tweets and filtered **2,302,924** epidemic tweets. Using a weak supervision approach, we trained models using four different epidemics of the silver standard data in a multi-classification setting and tested the models on publicly available large epidemic corpus. Our results demonstrate that weak supervision can be utilized in both class imbalance and multi-class setting and can efficiently classify several epidemic classes. Since this approach utilizes a noisy dataset curated using a labeling heuristic, the costs associated with manual annotation can be easily avoided.

# Appendix

## Data Collection details

Publicly available datasets - 2016 United States Presidential Election Tweet Ids (Littman et al. 2016), Online and Social Media Data As an Imperfect Continuous Panel Survey (Diaz et al. 2016), Eclipse tweet IDs (Summers), Hurricanes Harvey and Irma Tweet ids (Littman 2017b), Hurricane Florence Twitter Dataset (Phillips 2018), Hurricane Dorian Tweet IDs (Benjamin Rachunok et al. 2019), Hurricane Florence (Wrubel 2019), A Twitter Tale of Three Hurricanes: Harvey, Irma, and Maria (Alam et al. 2018), Puerto Rico tweets (Summers 2017), Tweet, but verify: epistemic study of information verification on Twitter (Zubiaga and Ji 2014), Hurricane Dorian Tweet IDs (Benjamin Rachunok et al. 2019), Hurricane Dorian Twitter Dataset (Phillips 2019), Distilling the Outcomes of Personal Experiences (Olteanu et al. 2017), Beyond the hashtags twitter data (Freelon), Climate change tweets ids (Littman and Wrubel 2019), Trump tweet IDs (Summers), Healthcare Tweet Ids (Littman 2019), 2018 U.S. Congressional Election Tweet Ids (Wrubel et al. 2018), News Outlet Tweet Ids (Littman et al. 2017b), #WomensMarch tweets (Ruest 2017b), U.S. government tweet ids (Littman et al. 2017a), End of term 2016 U.s. government twitter archive (Littman and Kerchner 2017), Nipsey Hussle tweets (Bergis Jules), Winter Olympics 2018 tweet ids (Littman 2018c), Dallas police shooting Twitter dataset (Phillips 2016), Charlottesville Tweet Ids (Littman 2018a), Twitter event datasets (2012-2016) (Zubiaga 2017), 115th U.S. Congress Tweet Ids (Littman 2017a), Immigration and travel ban tweet ids (Littman 2018b), Insight4news Irish news tweets (Poghosyan 2019), BlackLivesMatter tweets 2016 (Summers), 2020 US Presidential Election Tweet ID (Chen et al. 2021), Tweets to Donald Trump (Ruest 2017a).

Regex expression for filtering epidemic tweets: `"(?i:swine\s+flu|swineflu|h1n1|ebola|cholera|influenza|\bflu\b|yellow\s+fever|yellowfever|\bhiv\b|\b#aids\b|\#sars\b|\b#mers\b|\b#flu\b|\b#hiv\b)|\b#*AIDS\b|\bMERS\b|\bSARS\b"`

## Methods Technical details

For the classical models, the TF-IDF vectorizer was used to convert raw tweet text to TF-IDF features and return the document-term matrix which is sent to the model for training. We utilized LinearSVC for the SVM model and used the default parameters SVM. For the logistic regression we set "max_iter to 1000" and for the decision tree we set max_features to 'auto', criterion to "entropy" and max_depth to 150.

For the Transformer models' implementation, we utilized Simple Transformers (Rajapakse 2019) which seamlessly worked with the Natural Language Understanding (NLU) architectures made available by Hugging Face's Transformers models (Wolf et al. 2019). We employed early stopping techniques for the transformer model to avoid over fitting.

Table 4 presents the results from the Precision metric for each class for all the machine learning model experiments.

*Table 4*. Precision of machine learning models

| Epidemic Class | Classical Models | | | Deep Learning | |
|---|---|---|---|---|---|
| | DT | LR | **SVM** | **B** | BT |
| Cholera | 0.5372 | 0.9914 | **0.989** | **0.9959** | 0.9810 |
| Ebola | 0.6249 | 0.9617 | **0.9727** | **0.994** | 0.9974 |
| MERS | 0.3924 | 0.9016 | **0.8852** | **0.8118** | 0.7786 |
| Swine Flu | 0.7171 | 0.9977 | **0.9977** | **0.9997** | 0.9962 |
| non epidemic | 0.5974 | 0.9464 | **0.9469** | **0.9224** | 0.9193 |

Table 5 presents the results from the Recall metric for each class for all the machine learning model experiments.

*Table 5*. Recall of machine learning models

| Epidemic Class | Classical Models | | | Deep Learning | |
|---|---|---|---|---|---|
| | DT | LR | **SVM** | B | **BT** |
| Cholera | 0.6678 | 0.972 | **0.9781** | 0.9713 | **0.9835** |
| Ebola | 0.8737 | 0.9859 | **0.9868** | 0.9910 | **0.9874** |
| MERS | 0.9699 | 0.9701 | **0.9879** | 0.9965 | **0.9959** |
| Swine Flu | 0.6553 | 0.7314 | **0.7553** | 0.6995 | **0.6915** |
| non epidemic | 0.8237 | 0.9996 | **0.9998** | 0.9990 | **0.9993** |

Table 6 presents the results for the subset accuracy metric for all the machine learning model experiments.

*Table 6*. Accuracy of machine learning models

| Class | Accuracy |
|---|---|
| **SVM** | **0.9617** |
| LR | 0.9578 |
| DT | 0.6045 |
| B | 0.9549 |
| BT | 0.9532 |

Figures 1,2,3,4 depict the normalized confusion matrices for the top 4 models.

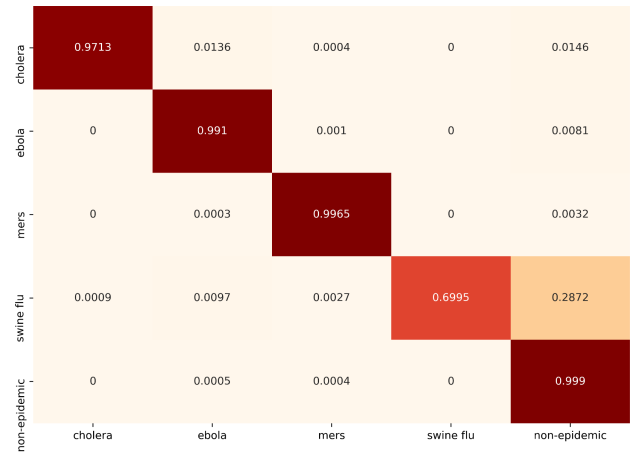

*Figure 1*. Confusion Matrix for BERT model

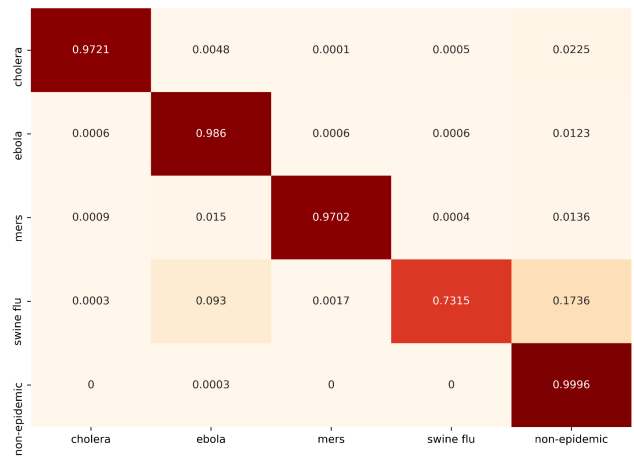

*Figure 3*. Confusion Matrix for Logistic Regression model

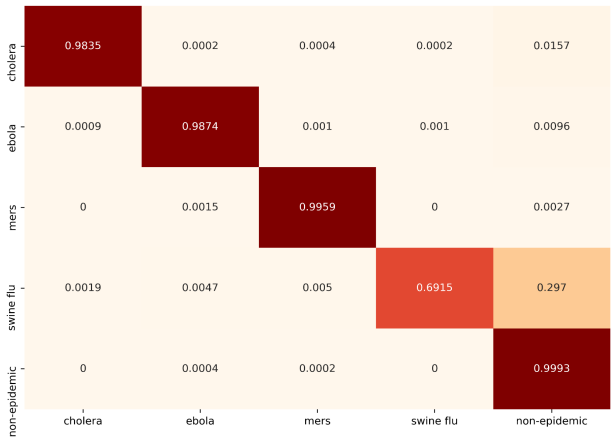

*Figure 2*. Confusion Matrix for BERTweet model

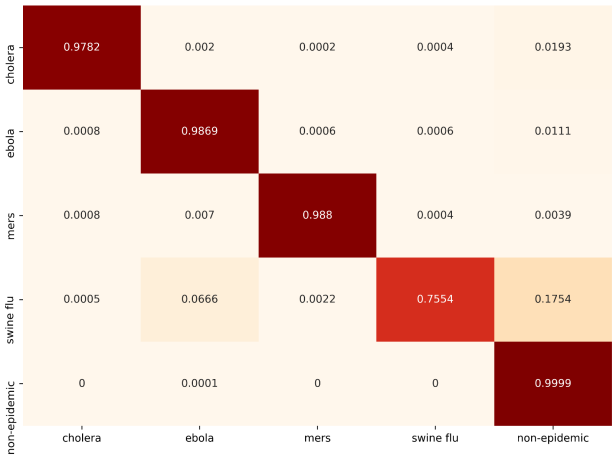

*Figure 4*. Confusion Matrix for SVM model